\documentclass[letterpaper]{article} % DO NOT CHANGE THIS
\usepackage{aaai24}
\usepackage{times}  % DO NOT CHANGE THIS
\usepackage{helvet}  % DO NOT CHANGE THIS
\usepackage{courier}  % DO NOT CHANGE THIS
\usepackage[hyphens]{url}  % DO NOT CHANGE THIS
\usepackage{graphicx} % DO NOT CHANGE THIS
\usepackage{natbib}  % DO NOT CHANGE THIS AND DO NOT ADD ANY OPTIONS TO IT
\usepackage{caption} % DO NOT CHANGE THIS AND DO NOT ADD ANY OPTIONS TO IT
\usepackage{algorithm}
\usepackage{algorithmic}
\usepackage{amsfonts}
\usepackage{amsmath}
\usepackage{booktabs}
\usepackage{multirow}
\usepackage{subcaption}

\usepackage{newfloat}
\usepackage{listings}
\DeclareCaptionStyle{ruled}{labelfont=normalfont,labelsep=colon,strut=off} % DO NOT CHANGE THIS
\floatstyle{ruled}
\newfloat{listing}{tb}{lst}{}
\floatname{listing}{Listing}
\title{Deep Reinforcement Learning and Mean-Variance Strategies for Responsible Portfolio Optimization}
\author{
    Fernando~Acero$^{1,2}$,
    Parisa~Zehtabi$^{1}$,
    Nicolas~Marchesotti$^{1}$,
    Michael~Cashmore$^{1}$,
    Daniele~Magazzeni$^{1}$,
    Manuela~Veloso$^{1}$
}
\affiliations{
    $^{1}$ J.P. Morgan AI Research \quad
    $^{2}$ University College London \quad \\
}

\begin{document}

\maketitle

\begin{abstract}
Portfolio optimization involves determining the optimal allocation of portfolio assets in order to maximize a given investment objective. Traditionally, some form of mean-variance optimization is used with the aim of maximizing returns while minimizing risk, however, more recently, deep reinforcement learning formulations have been explored. 
Increasingly, investors have demonstrated an interest in incorporating ESG objectives when making investment decisions, and modifications to the classical mean-variance optimization framework have been developed. In this work, we study the use of deep reinforcement learning for responsible portfolio optimization, by incorporating ESG states and objectives, and provide comparisons against modified mean-variance approaches. Our results show that deep reinforcement learning policies can provide competitive performance against mean-variance approaches for responsible portfolio allocation across additive and multiplicative utility functions of financial and ESG responsibility objectives. 
\end{abstract}

\section{Introduction}
Responsible investing, often also referred to as Socially Responsible Investments (SRI) or Ethical Investments, has become an increasingly relevant theme in the investment management industry. Clients and capital owners have increasingly started to consider the \emph{impact} of their investments as part of their investments due diligence decision-making process.
%Responsible investment is an increasingly relevant topic in finance, especially within the realm of asset management, where there is an increasing demand from clients to consider not only the financial returns of their investments but also their \emph{impact} on society. 
To this end, portfolio managers have begun to leverage Environmental, Social, and Governance (ESG) information in their investment process and portfolio offerings. %This has been further strengthened by increasing regulatory pressure surrounding ESG reporting as well as increased focus in sustainability and corporate responsibility by large corporations.
This has been further strengthened by increased focus on ESG reporting, sustainability and corporate responsibility by large corporations.

\begin{figure}
    \centering
    \includegraphics[width = 0.47\textwidth]{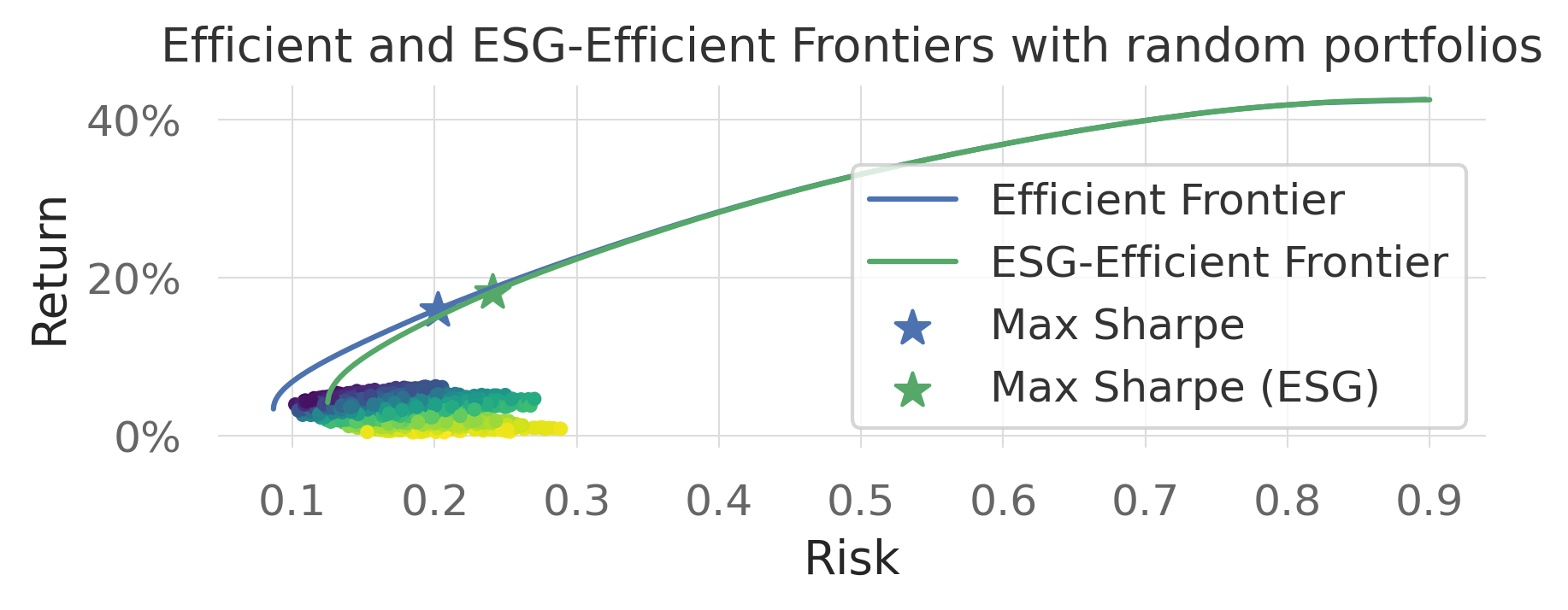}
    \caption{Efficient Frontiers for a sample of our evaluation data with and without constraints on portfolio ESG score. The ESG-Efficient Frontier always lies inside the Efficient Frontier \cite{pedersen2021responsible}.}
    \label{fig:esg_efficient_frontier}
\end{figure}

\textbf{Mean-variance optimization} (MVO), a cornerstone of Modern Portfolio Theory, is a framework to determine optimal capital allocation in portfolio optimization \cite{markowitz-mpt}. In its simplest form, it can be used to optimize expected returns for a desired risk level, or conversely to optimize risk for a desired return. The problem can be reformulated (as we detail in the Background section) to optimize for the Sharpe ratio or approximations thereof, resulting in the so-called tangency portfolio (highlighted in Figure \ref{fig:esg_efficient_frontier} as Max Sharpe). %, which is found at the tangent of the efficient frontier with the capital allocation line. 

Whilst hugely influential, various limitations of the original MVO approach have been previously identified, such as (i) high variability of optimal solutions to small changes in the inputs (expected return distributions) \cite{hurley2015note}, (ii) the use of standard deviation of expected returns as a measure of risk which penalizes returns below and above the expectation, as opposed to a semi-deviation which only accounts for downside risk or the use of higher-order moments \cite{harvey2010portfolio}, (iii) the lack of realistic factors impacting performance for portfolio managers such as transaction costs associated with frequent asset allocation adjustments \cite{lobo2007portfolio}. Most of the limitations are a consequence of MVO being formulated as a Quadratic Program (QP), which can easily be solved to certifiable convergence -- whereas incorporating modifications such as (ii) or (iii) may result in a general Nonlinear Program (NLP) which increases the difficulty of the optimization significantly and generally lacks optimality guarantees.

\textbf{Reinforcement Learning} (RL) is the machine learning paradigm for sequential decision-making. Recent literature has explored the use of Deep RL for multiple financial applications \cite{charpentier2021reinforcement, hambly2023recent, ozbayoglu2020deep}. 

With regards to portfolio optimization, \citet{ye2020reinforcement} propose a state-augmented RL framework for portfolio management which incorporates heterogeneous information for each asset as well as enhanced robustness against financial market uncertainty. \citet{jiang2017deep} propose a Deep RL framework for portfolio optimization which incorporates transaction costs and various neural network topologies, addressing the aforementioned limitation (ii) of MVO. \citet{sood2023deep} offer a comparative study between Deep RL and MVO for portfolio allocation. \citet{yu2019model} explore the use of model-based RL with adversarial generative models for dynamic portfolio optimization. More generally, \citet{liu2020finrl, liu2022finrl} provide frameworks and benchmarks for data-driven financial RL, evaluated in automated trading and portfolio management settings.

\textbf{ESG Portfolio Management}. The integration of ESG criteria into financial portfolios has been extensively discussed in the portfolio management literature \cite{henriksson2019integrating, branch2019guide, chen2021social}. The concept of the ESG-Efficient Frontier was introduced by \citet{pedersen2021responsible}, who studied the trade-off between financial and ESG performance of portfolios. In line with their results, we illustrate that trade-off for a sample of our evaluation data in Figure \ref{fig:esg_efficient_frontier}, which shows the ESG-Efficient Frontier for portfolios constrained to ESG scores at least 25\% better than the average ESG score across investable assets.
Lastly, we note that, although some authors have studied whether ESG objectives can drive or hurt portfolio returns \cite{halbritter2015wages, ouchen2022esg}, we consider these questions are beyond the scope of this work, as we are concerned with evaluating the potential for different ways to incorporate ESG objectives in Deep RL for portfolio optimization and its comparison to modified MVO approaches.

%In this paper, we study responsible portfolio optimization with ESG objectives using Deep RL and MVO for stocks currently included in the Dow Jones 30 Industrial Average index. We provide modifications to the traditional MVO approach to account for ESG objectives with additive utility, as well as RL formulations with additive and multiplicative ESG utility. Each method is implemented for Sharpe and Sortino objectives, where we propose a novel Differential Sortino ratio that we use for RL. Lastly, we provide a comprehensive evaluation of these methods in backtests against the financial and responsibility objectives, as well as additional performance metrics. 

%ESG efficient frontier and 3 types of investors: ESG-unaware, ESG-aware, ESG-motivated \cite{pedersen2021responsible}

In this paper, we study the use of Deep RL and MVO for responsible portfolio optimization with various forms of objectives. Our contributions are as follows:
\begin{itemize}
    \item We provide simple modifications to MVO formulations to accommodate for ESG objectives using additive utility, as a simplified version of previous work \cite{pedersen2021responsible} and a novel modification to an exact formulation \cite{cornuejols2006optimization}. 
    \item We suggest a formulation for Deep RL to incorporate ESG objectives, using either additive or multiplicative utility to incorporate responsibility objectives in the reward function. We also propose a modification of the Differential Sharpe ratio to yield a Differential Sortino ratio. 
    \item We perform a comprehensive evaluation of Deep RL and MVO strategies from the perspective of financial and responsibility objectives on recent evaluation data.
\end{itemize}

The remainder of this paper is structured as follows: we first provide relevant Background on MVO and RL, followed by our proposed approaches to Responsible Portfolio Optimization, discussing utility functions, MVO and RL. Then we describe our Experiments, followed by a discussion of Results. Lastly, we provide final remarks and directions for future work in our Conclusion.

\section{Background} \label{sec:background}
In this section we provide an overview of MVO and RL. 
\subsection{Mean-Variance Optimization}
Modern Portfolio Theory was originally introduced as a mathematical framework for assembling a portfolio of assets to minimize the level of risk such that the expected returns are above some threshold \cite{markowitz-mpt}. Formally, the optimization problem is stated as:
\begin{equation}
\label{eq:min_var}
    \min_{w} \, w^T \Sigma w
\end{equation}
s.t.
\begin{equation}
\label{eq:excess_ret}
    w^T \mu \geq \mu^{*}
\end{equation}
\begin{equation}
\label{eq:unity_w}
    w^T \mathbf{1} = 1 
\end{equation}
\begin{equation}
\label{eq:positive_w}
    w_i \geq 0
\end{equation}

where $w \in \mathbb{R}^N$ is the vector of weights denoting the distribution of $N$ assets in the portfolio, $\mu \in \mathbb{R}^N$ and $\Sigma \in \mathbb{R}^{N \times N}$ are beliefs about the future performance of the assets in the portfolio represented as expected mean and covariances respectively, and $\mu^{*}$ is some reference return usually taken as the risk-free return available in the market (e.g. short-term government securities). Constraint \ref{eq:excess_ret} enforces returns above a baseline, \ref{eq:unity_w} implies the full portfolio amount has to be invested (without leverage), and \ref{eq:positive_w} implies the portfolio is long-only (i.e. no short-selling of assets is allowed).  

Practically, the mean-variance optimization problem is often reformulated to solve for the tangency portfolio, where the objective is to maximize the expected \textit{Sharpe ratio} of the portfolio (excess returns over risk-free rate per unit of increase in risk), which is stated as:
\begin{equation}
\label{eq:max_sharpe}
    \max_{w} \, \frac{\mu^T w - r_f}{\sqrt{w^T \Sigma w}}
\end{equation}
subject to \ref{eq:unity_w} and \ref{eq:positive_w}, where $r_f$ is the return of the risk-free asset. 

The objective in \ref{eq:max_sharpe} is not convex, and several approaches have been used to solve it in spite of this, such as the bisection method, the Dinkelbach transform \cite{dinkelbach1967nonlinear}, or the Schaible transform \cite{schaible1974parameter}. However, those are iterative approaches. Alternatively, the following transform yields the optimal solution to \ref{eq:max_sharpe} directly under mild assumptions as shown by \citet{cornuejols2006optimization}: let $\hat{\mu}$ be given by $\hat{\mu}_i = \mu_i - r_f$ for each asset $i$, and let $y = \kappa w$ with $\kappa = \frac{1}{\hat{\mu}^T w}$. Then, $\sqrt{w^T \Sigma w} = \frac{1}{\kappa} \sqrt{y^T \Sigma y}$, and $\frac{1}{\hat{\mu}^T w} \Leftrightarrow \hat{\mu}^T y = 1$ given $\frac{y}{\kappa}=w$. Thus, \ref{eq:max_sharpe} is equivalent to the quadratic program (QP):
\begin{equation}
    \label{eq:max_sharpe_transformed}
    \min_y \, y^T \Sigma y
\end{equation}
s.t.
\begin{equation}
    \label{eq:max_sharpe_transformed_ymu}
    \hat{\mu}^T y = 1
\end{equation}
\begin{equation}
    \label{eq:max_sharpe_transformed_yk}
    y^T \mathbf{1} = \kappa , \kappa > 0
\end{equation}
\begin{equation}
    \label{eq:max_sharpe_transformed_y}
    y_i \geq 0
\end{equation}

\subsection{Reinforcement Learning}

Reinforcement Learning lies at the intersection of machine learning and optimal control, and is concerned with how an agent ought to take actions in an environment in order to maximize expected future rewards \cite{sutton_reinforcement_2018}. The sequential decision making problem is often formulated as a Markov Decision Process (MDP), defined as a 4-tuple $\langle \mathcal{S}, \mathcal{A}, \mathcal{T}, \mathcal{R} \rangle$ where $\mathcal{S}$ is the state space, $\mathcal{A}$ is the action space, $\mathcal{T}$ is the set of transition probabilities from states $s_t$ to $s_{t+1}$ when taking action $a_t$ (i.e. the dynamics), and $\mathcal{R}$ is the reward function. A discount factor $\gamma$ is typically introduced to discount future rewards. We refer to a policy $\pi:\mathcal{S} \rightarrow \mathcal{A}$ parametrized by $\theta$ as $\pi_{\theta}$.

Multiple taxonomies of RL algorithms have been proposed. A possible categorization is that of model-based or model-free algorithms, where model-based approaches solve planning problems by making use of a given model of the MDP or learning one from experience, whereas model-free approaches aim to learn a reactive control policy that directly maps state observations to actions. Alternatively, algorithms may be categorized as value-based and/or policy-based, depending on whether a policy (actor) and/or a value function (critic) is learned. Deep RL leverages neural networks as general function approximators and deep learning techniques for the policy or value function of RL algorithms. 

Actor-critic algorithms are a popular approach amongst practitioners as they combine the advantage of policy-based methods and value-based methods: a parametric policy is explicitly learned via some estimate of the policy gradient, which uses advantage estimates that rely on a value function that is learned concurrently in order to reduce variance of the policy gradient estimates. Learning an explicit policy is useful as it can be used at test time directly, without performing any kind of optimization or querying of the gradient of a value function (in continuous state-action spaces) or the maximum of q-value function (in discrete state-action spaces). Policy gradient algorithms \cite{SuttonPolicy} can be used for discrete or continuous action spaces, and usually optimize some form of the policy gradient:
\begin{equation}
    \nabla_{\theta} \mathbb{E} \left[\sum_{t=0}^{T} r_t \right] \approx \mathbb{E} \left[ \sum_{t=0}^{T} \Psi_t \nabla_{\theta} \log \pi_{\theta}(a_t | s_t) \right]
    \label{eq:policygradient}
\end{equation}
where $\Psi_t$ can be the discounted returns of the trajectory or temporal-difference residual, but typically involves some advantage estimate in actor-critic algorithms \cite{Schulmanetal_ICLR2016}. 
A popular policy gradient algorithm is Proximal Policy Optimization (PPO) \cite{PPOSchulman}, which uses a clipped surrogate objective to \eqref{eq:policygradient} providing a lower bound on the unclipped objective. This approach avoids large policy updates which could be detrimental, and usually provides robust performance. For these reasons, we use PPO as our RL algorithm in an actor-critic fashion by learning a value function to produce advantage estimates using Generalized Advantage Estimation (GAE) to address the bias-variance trade-off in advantage estimates \cite{Schulmanetal_ICLR2016}.

%%% useful links
% https://icaps23.icaps-conference.org/papers/finplan/FinPlan23_paper_4.pdf
% https://arxiv.org/pdf/2201.01227.pdf
% https://arxiv.org/pdf/2308.15135.pdf
% https://www.northinfo.com/documents/601.pdf
% https://arxiv.org/pdf/1906.01509.pdf

% https://www.sciencedirect.com/science/article/abs/pii/S1062940822001280
% https://arxiv.org/ftp/arxiv/papers/2207/2207.02134.pdf

\section{Responsible Portfolio Optimization}
\label{sec:responsible_port_opt}

We elaborate on different ways in which responsibility objectives can be introduced in portfolio optimization, followed by the problem formulations and methodologies we use to perform portfolio optimization with ESG objectives, namely mean-variance and RL approaches. %We include variants for optimizing the Sharpe ratio and the (approximate) Sortino ratio for each approach, and include ESG, E, S, or G portfolio scores as additional objectives. 

\subsection{Incorporating Responsibility Objectives}
Portfolio managers follow various methodologies to responsible investing. \emph{Integration} approaches explicitly and systematically include ESG issues in the investment decisions, \emph{screening} methods apply filters based on rules as a pre- or post-processing, and \emph{thematic} approaches tend to seek to contribute to a specific environmental or social outcome. 
Methods that do not incorporate responsibility objectives in the optimization problem are generally suboptimal as they modify inputs or solutions
%The methodology used by portfolio managers for incorporating responsible investing objectives has historically involved methods such as asset screening, where the assets in the investable universe are restricted prior to portfolio optimization or the asset allocation resulting from the optimization process is adjusted a posteriori according to responsibility criteria. These methods are generally suboptimal as they pre- or post-process the inputs or solutions, instead of incorporating responsibility objectives in the optimization problem
\cite{verheyden2016esg, jin2022esg, qi2020imposing}. 

Integration methods are the more principled approach, as they introduce a responsibility term in the objective function. Regardless of the financial objective used for portfolio optimization (e.g. Sharpe, Sortino, Information ratios), incorporating responsibility objectives such as ESG scores in the optimization process involves devising an appropriate utility function to reflect the motivation of an investor. Consider a financial objective $u_{\text{financial}}$ and a responsibility objective $u_{\text{responsible}}$. A general form for a utility function is $\mathcal{U} = f(u_{\text{financial}}, u_{\text{responsible}}) $. A simple utility function of the additive form:

\begin{equation}
    \label{eq:utility_fn_additive}
    \mathcal{U}_{\text{additive}} = u_{\text{financial}} + u_{\text{responsible}}
\end{equation}
is a possible candidate for representing the objective of a responsible investor. Objectives of this form are studied by \citet{pedersen2021responsible}, who use a bi-level approach to solve for objectives of this form for ESG-motivated investors. Their results demonstrate a trade-off between ESG and Sharpe ratio of the portfolio, which we partly replicate in Figure \ref{fig:esg_efficient_frontier}. We interpret this as the consequence of Pareto-optimal solutions in a bi-objective optimization problem with an additive objective function. These objectives are amenable to traditional mean-variance approaches as the utility function can remain convex. We note how an alternative approach is to introduce the responsibility objective in an multiplicative way, yielding utility functions of the form:

\begin{equation}
    \label{eq:utility_fn_multiplicative}
    \mathcal{U}_{\text{multiplicative}} = u_{\text{financial}} \cdot u_{\text{responsible}}
\end{equation}
which may be desirable as it effectively scales the financial utility attained by a portfolio by a responsibility term. This objective is not convex and thus cannot be optimized for using QP formulations, which are used for mean-variance optimization, further motivating the use of RL for responsible portfolio optimization.

\subsection{Mean-Variance Optimization with ESG objectives}
Our formulation for MVO leverages the transformation proposed by \citet{cornuejols2006optimization} described in the Background section, where the optimization objective is given by \ref{eq:max_sharpe_transformed}. 

We formulate MVO problems that incorporate ESG terms into the objective function in an additive way in order to preserve the convexity of the objective to be able to use a QP solver. We denote the vector of responsiblity scores for the assets in the portfolio as $s \in \mathbb{R}^N$, with different versions of this being $ s|_{ESG} , s|_{E} , s|_{S}, s|_{G}$ depending on which responsibility score is considered. In order to account for varying levels of mean responsibility scores through time for assets in the portfolio, we propose to incorporate the objective $u_{\text{responsible}}$ based on the ratio of the score of our portfolio (the one using $y$ weights for asset allocation) and the score of a uniform-allocation portfolio (the one that uses a uniform vector $u\in \mathbb{R}^N$ with entries $u_i = (1/N)  \forall i \in \{1, ..., N\}$ for asset allocation). The use of the ratio of the portfolio score over the uniform-allocation score is motivated by \citet{pedersen2021responsible}. We introduce a scaling coefficient $\alpha$ to control the desired investor sensitivity to the responsibility performance. Thus, our optimization formulation is:

%\begin{equation}
%    \min_y y^T \Sigma y - \alpha \frac{\sum_{1}^{N} (\frac{1}{N} y_i \cdot s_i)}{\sum_{i}^{N} (\frac{1}{N} u_i \cdot s_i)} 
%\end{equation}

\begin{equation}
    \label{eq:max_sharpe_transformed_responsible}
    \min_y \, y^T \Sigma y - \alpha \frac{y^T s}{u^T s} 
\end{equation}
subject to \ref{eq:max_sharpe_transformed_ymu}, \ref{eq:max_sharpe_transformed_yk}, \ref{eq:max_sharpe_transformed_y}.
Note the negative sign in front of the responsibility score term $\frac{y^T s}{u^T s} $, which is needed within the minimization problem in order to maximize the ratio of responsibility score of the portfolio over the uniform-allocation portfolio. We note how $s$ in \ref{eq:max_sharpe_transformed_responsible} can correspond to $ s|_{ESG} , s|_{E} , s|_{S}, s|_{G}$ or linear combinations thereof depending on the investor preference. An alternative approach, based on a relaxation of the Sharpe ratio objective which is frequently used is defined as:
\begin{equation}
    \label{eq:relaxed_max_sharpe_transformed_responsible}
    \max_w \, \mu^T w -\lambda w^T \Sigma w + \alpha \frac{w^T s}{u^T s} 
\end{equation}
subject to \ref{eq:unity_w} and \ref{eq:positive_w}, where $\lambda$ is a parameter that controls the risk-aversion of the investor and the responsibility objective is the same as in \ref{eq:max_sharpe_transformed_responsible}. This is a simplified form of the bi-level optimization of \citet{pedersen2021responsible}. 

Following the material discussed in the Background section, it can be seen that the formulation in \ref{eq:max_sharpe_transformed_responsible} provides a maximization of a linear combination of the Sharpe ratio and the responsibility score of the portfolio, and \ref{eq:relaxed_max_sharpe_transformed_responsible} a relaxed approximation thereof. The formulations in \ref{eq:max_sharpe_transformed_responsible} or \ref{eq:relaxed_max_sharpe_transformed_responsible} may also be used to optimize for a combination of the (approximate) Sortino ratio and responsibility score, if a semicovariance matrix is used instead, instead of the sample covariance matrix $\Sigma$, yielding a mean-semivariance optimization. To do so, we follow the approximation proposed by \citet{estrada2008mean}, which estimates the semicovariance matrix as:
\begin{equation}
    \Sigma^{-} \approx \frac{1}{N} \sum_{i=1}^{N} \sum_{j=1}^{N} \min (r_i , r_f) \cdot \min (r_j , r_f)
\end{equation}
where we use $-$ in the superscript to denote this is the semicovariance for returns below the benchmark (i.e. representing downside risk), $r_i, r_j$ are returns of portfolio assets, and the risk-free return $r_f$ is the benchmark return.

\begin{figure*}[]
\centering
\hspace*{7.5em}
  \begin{subfigure}[c]{0.475\textwidth}
  \centering
    \includegraphics[width=\linewidth]{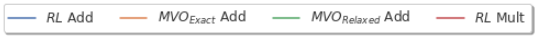}
  \end{subfigure}
  \hfill %%
  \newline
  \begin{subfigure}[c]{0.425\textwidth}
    \includegraphics[width=\linewidth]{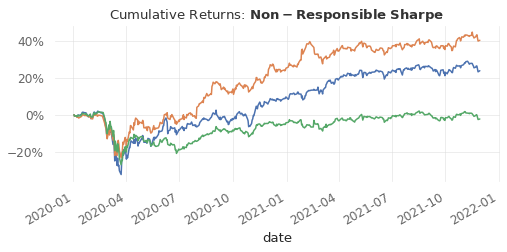}
    %\caption{Picture 1}
    \label{fig:results1Sharpe}
  \end{subfigure}
  %\hfill %%
  \begin{subfigure}[c]{0.425\textwidth}
    \includegraphics[width=\linewidth]{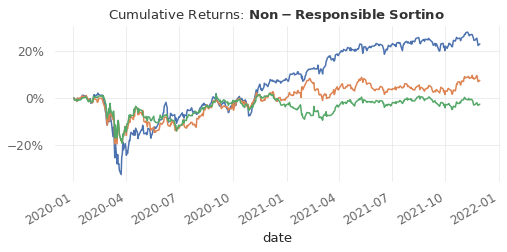}
    %\caption{Picture 2}
    \label{fig:results1Sortino}
  \end{subfigure}
    %\hfill %%
  \begin{subfigure}[c]{0.425\textwidth}
    \includegraphics[width=\linewidth]{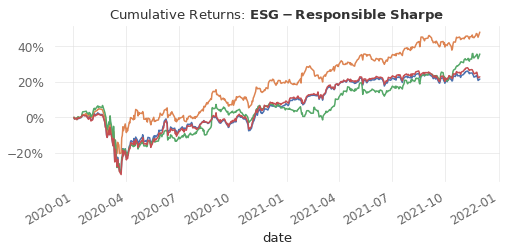}
    %\caption{Picture 1}
    \label{fig:results2Sharpe}
  \end{subfigure}
  %\hfill %%
  \begin{subfigure}[c]{0.425\textwidth}
    \includegraphics[width=\linewidth]{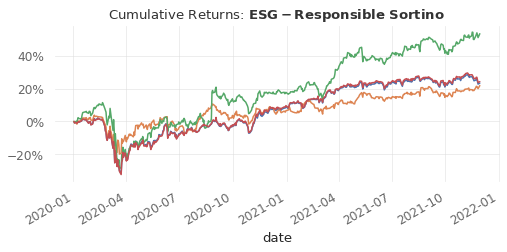}
    %\caption{Picture 2}
    \label{fig:results2Sortino}
  \end{subfigure}
    %\hfill %%
  \begin{subfigure}[c]{0.425\textwidth}
    \includegraphics[width=\linewidth]{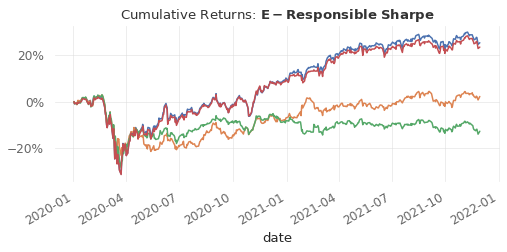}
    %\caption{Picture 1}
    \label{fig:results3Sharpe}
  \end{subfigure}
  %\hfill %%
  \begin{subfigure}[c]{0.425\textwidth}
    \includegraphics[width=\linewidth]{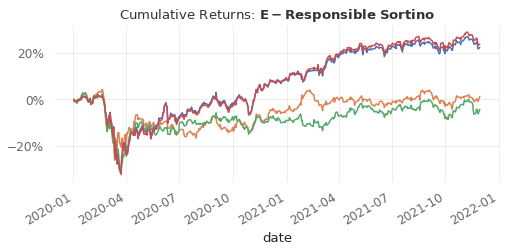}
    %\caption{Picture 2}
    \label{fig:results3Sortino}
  \end{subfigure}
  \begin{subfigure}[c]{0.425\textwidth}
    \includegraphics[width=\linewidth]{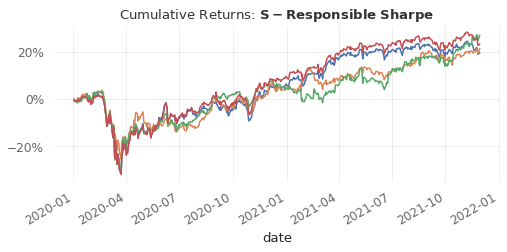}
    %\caption{Picture 1}
    \label{fig:results4Sharpe}
  \end{subfigure}
  %\hfill %%
  \begin{subfigure}[c]{0.425\textwidth}
    \includegraphics[width=\linewidth]{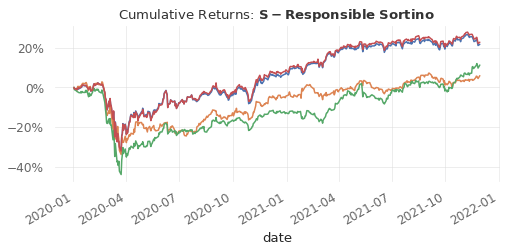}
    %\caption{Picture 2}
    \label{fig:results4Sortino}
  \end{subfigure}
  \begin{subfigure}[c]{0.425\textwidth}
    \includegraphics[width=\linewidth]{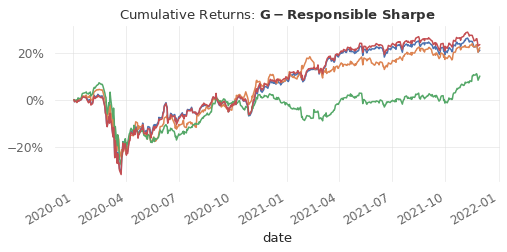}
    %\caption{Picture 1}
    \label{fig:results5Sharpe}
  \end{subfigure}
  %\hfill %%
  \begin{subfigure}[c]{0.425\textwidth}
    \includegraphics[width=\linewidth]{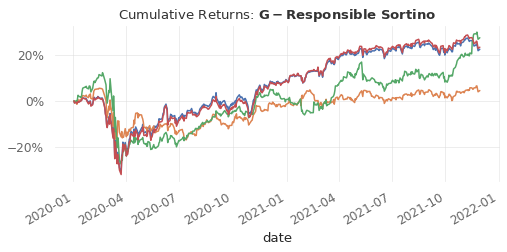}
    %\caption{Picture 2}
    \label{fig:results5Sortino}
  \end{subfigure}
  \caption{Cumulative financial returns for RL, $\text{MVO}_{\text{Exact}}$, $\text{MVO}_{\text{Relaxed}}$ for various utility functions in evaluation period (for non-responsible results, \emph{Add} in legend is redundant, due to zero responsibility objective).}
  \label{fig:results_all}
\end{figure*}

\subsection{Deep Reinforcement Learning with ESG objectives}
In order to train RL policies for portfolio optimization, we formulate MDPs similar to that of \citet{sood2023deep}, building upon the implementation of \citet{liu2022finrl}. The state space includes the same $\mu$ and $\Sigma$ as those used for mean-variance optimization, as well as the $\mu$ values corresponding to a lookback period of $T$ days. Additionally, we include terms based on technical analysis indicators which provide the RL agent with additional features that may be useful nonlinear features of the price dynamics of an asset. The technical indicators used are Simple Moving Average (SMA), Moving Average Convergence Divergence (MACD), Relative Strength Index (RSI), Bollinger Bands, Commodity Channel Index (CCI), and Average Directorial Index (ADX), similar to prior work \cite{liu2022finrl}. Lastly, the state space also includes responsibility scores $ s|_{ESG} , s|_{E} , s|_{S}, s|_{G}$. The action space is the set of possible portfolio weights considering constraints \ref{eq:positive_w} and \ref{eq:unity_w}. 

The definition of the reward functions suitable for financial trading and portfolio optimization applications is not trivial, due to the fact that objectives such as the Sharpe or Sortino ratios are computed using moments estimated from samples over time, whereas RL is formulated as a sequential decision making problem and thus requires reward functions that reflect the sequential (step-by-step) nature of the process, as well as the fact that in RL the returns are additive on (discounted) rewards, whereas the Sharpe or Sortino ratios are not additive functions over time. To account for this, \citet{moody1998performance} proposed the use of a \emph{Differential Sharpe ratio}. This approach, which we consider more correct for RL than the naive (non-differential) approach, has been followed by some authors \cite{moody1998reinforcement, sood2023deep}, but not all previous work \cite{liu2022finrl}. We use the Differential Sharpe ratio given by:
\begin{equation}
\label{eq:differential_sharpe}
    D_t := \frac{\partial S_t}{\partial \eta} = \frac{B_{t-1} \Delta A_t  - \frac{1}{2} A_{t-1} \Delta B_t }{ (B_{t-1} - A^2_{t-1})^{3/2} }
\end{equation}

\begin{align*}
    A_t = A_{t-1} + \eta \Delta A_t   &&  \Delta A_t = R_t - A_{t-1} \\
     B_t = B_{t-1} + \eta \Delta B_t  &&  \Delta B_t = R^2_t - B_{t-1}
\end{align*}
for timescale $\eta$, derived from a first-order Taylor expansion in $\eta$ \cite{moody1998performance}. Note that in \ref{eq:differential_sharpe}, $R_t$ denotes the financial return of the portfolio (not to be mistaken for the returns in RL terminology, i.e. the discounted sum of rewards in the MDP). Inspired by the approximation proposed by \citet{estrada2008mean} for the semicovariance, we define the \emph{Differential Sortino ratio} just as in \ref{eq:differential_sharpe} except that we let $\Delta B_t = \min(R_t, r_f) \cdot \min(R_t, r_f)  - B_{t-1}$ where $r_f$ refers to the benchmark return (risk-free return). Our reward functions are thus the Differential Sharpe or Differential Sortino ratios, depending on the objective chosen for $u_{\text{financial}}$. Additionally, we incorporate $u_{\text{responsible}}$ for ESG, E, S, G objectives with additive or multiplicative utilities as chosen.

\section{Experiments}
Our experiments use the same environment for both portfolio optimization strategies, henceforth referred to as MVO and RL. Note that RL requires a training phase before rolling out the learned policy during evaluation, whereas MVO solves the optimization ad-hoc during evaluation. We refer to \ref{eq:max_sharpe_transformed_responsible} as $\text{MVO}_{\text{Exact}}$ and to \ref{eq:relaxed_max_sharpe_transformed_responsible} as $\text{MVO}_{\text{Relaxed}}$. We use data for 29 of the 30 stocks currently present in the Dow Jones Industrial Average as our investable universe (due to ESG data availability issues for one stock). We leverage public daily price data from \emph{Yahoo! Finance}, and the corresponding monthly ESG, E, S, and G scores provided by \emph{Sustainalytics}. 

To evaluate financial performance, we report annualized returns for RL, $\text{MVO}_{\text{Exact}}$, and $\text{MVO}_{\text{Relaxed}}$ for $u_{\text{financial}}$ being $u_{\text{Sharpe}}$ and $u_{\text{Sortino}}$. To evaluate responsibility performance, we define the performance ratio $p_r$ for each responsibility objective:

\begin{equation}
    p_r|_i = \left( \frac{ w^T s|_{i} }{ u^T s|_{i} } -1 \right) \quad \forall i \in \left[ESG, E, S, G\right]
\end{equation}
which is positive for responsibility performance above the uniform-allocation portfolio and negative for performance below it. 
Our training data spans from 2014-01-01 to 2019-11-30, and the evaluation data from 2020-01-01 to 2021-11-30. We set $r_f = 0$ during evaluation as this was approximately the rate of return for short-term US government bonds during the evaluation period. For RL, we use $\eta = \frac{1}{252}$ based on trading days per year, and PPO hyperparameters are default from \citet{raffin2021stable}. For all strategies, we use $\alpha = 0.1$, and for $\text{MVO}_{\text{Relaxed}}$ we use $\lambda = 10$.

We perform experiments with $u_{\text{financial}} = u_{\text{Sharpe}}$ and $u_{\text{financial}} = u_{\text{Sortino}}$ for both MVO approaches and for RL policies. Our experiments include a non-responsible setting (i.e. $\mathcal{U} = f(u_{\text{financial}})$), and responsible settings (i.e. $\mathcal{U} = f(u_{\text{financial}}, u_{\text{responsible}})$) with ESG, E, S, and G objectives for $u_{\text{responsible}}$. As previously discussed, for MVO approaches we only incorporate responsibility terms in an additive manner (i.e. $\mathcal{U} = \mathcal{U}_{\text{additive}}$) due to the convexity requirement, whereas for RL we use modified objectives that incorporate responsibility terms additively or multiplicatively (i.e. $\mathcal{U} = \mathcal{U}_{\text{multiplicative}}$) to the utility function as there are no convexity requirements for the rewards of an MDP in Deep RL. 

We perform training and evaluation using a Linux machine with an 8-core Intel Xeon Platinum CPU and a 16GB Nvidia Tesla T4 GPU. Training for RL policies across all configurations required 6 hours of wall-clock time.

\begin{table}[]
\centering
\caption{Annualized financial returns for RL, $\text{MVO}_{\text{Exact}}$, $\text{MVO}_{\text{Relaxed}}$ for various utility functions in evaluation period.}
\label{tab:results_fin}
\begin{tabular}{@{}cccccc@{}}
\toprule
\multicolumn{3}{c}{Utility Function}                                                             & RL & $\text{MVO}_{\text{Exact}}$ & $\text{MVO}_{\text{Relaxed}}$ \\ \midrule
\multicolumn{1}{c|}{\multirow{9}{*}{\rotatebox[origin=c]{90}{$u_{\text{financial}} = \text{Sharpe}$}}}  & \multicolumn{2}{c}{$u_{\text{Sharpe}}$}                         &  11.77 \% &   19.29 \% &      -1.2 \%           \\ \cmidrule(lr){2-3}
\multicolumn{1}{c|}{}                         & \multicolumn{2}{c}{$u_{\text{Sharpe}} + u_{\text{ESG}}$}                         &  10.76 \% &   22.77 \% &     17.32 \%           \\
\multicolumn{1}{c|}{}                         & \multicolumn{2}{c}{$u_{\text{Sharpe}} \cdot u_{\text{ESG}}$} &  11.37 \% &        - &          -           \\ \cmidrule(lr){2-3}
\multicolumn{1}{c|}{}                         & \multicolumn{2}{c}{$u_{\text{Sharpe}} + u_{\text{E}}$}  &  12.53 \% &    1.16 \% &     -6.86 \%           \\
\multicolumn{1}{c|}{}                         & \multicolumn{2}{c}{$u_{\text{Sharpe}} \cdot u_{\text{E}}$} &  11.65 \% &        - &          -           \\ \cmidrule(lr){2-3}
\multicolumn{1}{c|}{}                         & \multicolumn{2}{c}{$u_{\text{Sharpe}} + u_{\text{S}}$}  &    9.9 \% &   10.72 \% &     13.42 \%           \\
\multicolumn{1}{c|}{}                         & \multicolumn{2}{c}{$u_{\text{Sharpe}} \cdot u_{\text{S}}$} &  11.63 \% &        - &          -           \\ \cmidrule(lr){2-3}
\multicolumn{1}{c|}{}                         & \multicolumn{2}{c}{$u_{\text{Sharpe}} + u_{\text{G}}$}  &  10.47 \% &   11.01 \% &      5.12 \%           \\
\multicolumn{1}{c|}{}                         & \multicolumn{2}{c}{$u_{\text{Sharpe}} \cdot u_{\text{G}}$} &  11.62 \% &        - &          -           \\ \cmidrule(r){1-3}
\multicolumn{1}{c|}{\multirow{9}{*}{\rotatebox[origin=c]{90}{$u_{\text{financial}} = \text{Sortino}$}}} & \multicolumn{2}{c}{$u_{\text{Sortino}}$}                  &  11.45 \% &    3.87 \% &     -1.39 \%           \\ \cmidrule(lr){2-3}
\multicolumn{1}{c|}{}                         & \multicolumn{2}{c}{$u_{\text{Sortino}} + u_{\text{ESG}}$}   &   11.6 \% &   10.89 \% &     25.08 \%           \\
\multicolumn{1}{c|}{}                         & \multicolumn{2}{c}{$u_{\text{Sortino}} \cdot u_{\text{ESG}}$}  &  12.03 \% &        - &          -           \\ \cmidrule(lr){2-3}
\multicolumn{1}{c|}{}                         & \multicolumn{2}{c}{$u_{\text{Sortino}} + u_{\text{E}}$}   &  11.05 \% &    0.65 \% &     -2.21 \%           \\
\multicolumn{1}{c|}{}                         & \multicolumn{2}{c}{$u_{\text{Sortino}} \cdot u_{\text{E}}$}  &  11.76 \% &        - &          -           \\ \cmidrule(lr){2-3}
\multicolumn{1}{c|}{}                         & \multicolumn{2}{c}{$u_{\text{Sortino}} + u_{\text{S}}$}   &   10.8 \% &    3.05 \% &      5.73 \%          \\
\multicolumn{1}{c|}{}                         & \multicolumn{2}{c}{$u_{\text{Sortino}} \cdot u_{\text{S}}$}  &   11.3 \% &        - &          -          \\ \cmidrule(lr){2-3}
\multicolumn{1}{c|}{}                         & \multicolumn{2}{c}{$u_{\text{Sortino}} + u_{\text{G}}$}   &  11.14 \% &    2.33 \% &     13.49 \%           \\
\multicolumn{1}{c|}{}                         & \multicolumn{2}{c}{$u_{\text{Sortino}} \cdot u_{\text{G}}$}  &  11.56 \% &        - &          -          \\ \bottomrule
\end{tabular}
\end{table}

\begin{table}[]
\centering
\caption{Average $p_r|_i$ for RL, $\text{MVO}_{\text{Exact}}$, $\text{MVO}_{\text{Relaxed}}$ for $i \in \left[ESG, E, S, G\right]$ in evaluation period. For non-responsible objectives we report $p_r|_{ESG}$.}
\label{tab:results_resp}
\begin{tabular}{@{}cccccc@{}}
\toprule
\multicolumn{3}{c}{Utility Function}                                                             & RL & $\text{MVO}_{\text{Exact}}$ & $\text{MVO}_{\text{Relaxed}}$ \\ \midrule
\multicolumn{1}{c|}{\multirow{9}{*}{\rotatebox[origin=c]{90}{$u_{\text{financial}} = \text{Sharpe}$}}}  & \multicolumn{2}{c}{$u_{\text{Sharpe}}$}                         &  -0.13 \% &    0.69 \% &      0.04 \%           \\ \cmidrule(lr){2-3}
\multicolumn{1}{c|}{}                         & \multicolumn{2}{c}{$u_{\text{Sharpe}} + u_{\text{ESG}}$}                         &    0.9 \% &    4.29 \% &     11.69 \%          \\
\multicolumn{1}{c|}{}                         & \multicolumn{2}{c}{$u_{\text{Sharpe}} \cdot u_{\text{ESG}}$} &  -0.03 \% &        - &          -           \\ \cmidrule(lr){2-3}
\multicolumn{1}{c|}{}                         & \multicolumn{2}{c}{$u_{\text{Sharpe}} + u_{\text{E}}$}  &   0.54 \% &    1.61 \% &      2.34 \%           \\
\multicolumn{1}{c|}{}                         & \multicolumn{2}{c}{$u_{\text{Sharpe}} \cdot u_{\text{E}}$} &   0.08 \% &        - &          -           \\ \cmidrule(lr){2-3}
\multicolumn{1}{c|}{}                         & \multicolumn{2}{c}{$u_{\text{Sharpe}} + u_{\text{S}}$}  &    0.7 \% &    0.53 \% &       4.7 \%           \\
\multicolumn{1}{c|}{}                         & \multicolumn{2}{c}{$u_{\text{Sharpe}} \cdot u_{\text{S}}$} &  -0.05 \% &        - &          -          \\ \cmidrule(lr){2-3}
\multicolumn{1}{c|}{}                         & \multicolumn{2}{c}{$u_{\text{Sharpe}} + u_{\text{G}}$}  &   0.11 \% &    0.99 \% &      2.16 \%          \\
\multicolumn{1}{c|}{}                         & \multicolumn{2}{c}{$u_{\text{Sharpe}} \cdot u_{\text{G}}$} &   0.03 \% &        - &          -           \\ \cmidrule(r){1-3}
\multicolumn{1}{c|}{\multirow{9}{*}{\rotatebox[origin=c]{90}{$u_{\text{financial}} = \text{Sortino}$}}} & \multicolumn{2}{c}{$u_{\text{Sortino}}$}                  &  -0.16 \% &    1.07 \% &       0.5 \%           \\ \cmidrule(lr){2-3}
\multicolumn{1}{c|}{}                         & \multicolumn{2}{c}{$u_{\text{financial}} + u_{\text{ESG}}$}   &   0.38 \% &    5.25 \% &     13.35 \%          \\
\multicolumn{1}{c|}{}                         & \multicolumn{2}{c}{$u_{\text{Sortino}} \cdot u_{\text{ESG}}$}  &  -0.16 \% &        - &          -         \\ \cmidrule(lr){2-3}
\multicolumn{1}{c|}{}                         & \multicolumn{2}{c}{$u_{\text{Sortino}} + u_{\text{E}}$}   &    0.1 \% &    2.14 \% &      3.02 \%            \\
\multicolumn{1}{c|}{}                         & \multicolumn{2}{c}{$u_{\text{Sortino}} \cdot u_{\text{E}}$}  &   0.03 \% &        - &          -            \\ \cmidrule(lr){2-3}
\multicolumn{1}{c|}{}                         & \multicolumn{2}{c}{$u_{\text{Sortino}} + u_{\text{S}}$}   &   0.37 \% &    0.87 \% &       5.7 \%          \\
\multicolumn{1}{c|}{}                         & \multicolumn{2}{c}{$u_{\text{Sortino}} \cdot u_{\text{S}}$}  &   0.05 \% &        - &          -          \\ \cmidrule(lr){2-3}
\multicolumn{1}{c|}{}                         & \multicolumn{2}{c}{$u_{\text{Sortino}} + u_{\text{G}}$}   &    0.1 \% &    1.17 \% &      2.74 \%            \\
\multicolumn{1}{c|}{}                         & \multicolumn{2}{c}{$u_{\text{Sortino}} \cdot u_{\text{G}}$}  &  -0.07 \% &        - &          -          \\ \bottomrule
\end{tabular}
\end{table}

\section{Results}

Firstly, we note that the evaluation period spans nearly 2 years from January 2020, and thus is heavily affected by the effects of Covid-19 on financial markets. We purposefully use such a volatile regime to evaluate the performance of RL policies with data far from the training distribution.  

\begin{figure*}[]
\hspace*{12.5em}
  \begin{subfigure}[c]{0.475\textwidth}
  \centering
    \includegraphics[width=\linewidth]{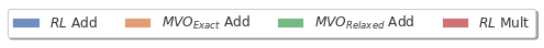}
  \end{subfigure}
  \hfill %%
  \newline
  \begin{subfigure}[c]{0.49\textwidth}
    \includegraphics[width=\linewidth]{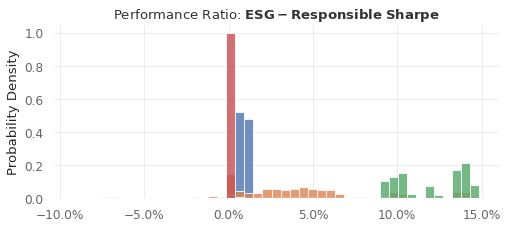}
    %\caption{Picture 1}
    \label{fig:prSharpe}
  \end{subfigure}
  %\hfill %%
  \begin{subfigure}[c]{0.49\textwidth}
    \includegraphics[width=\linewidth]{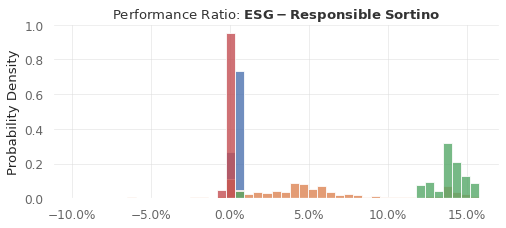}
    %\caption{Picture 2}
    \label{fig:prSortino}
  \end{subfigure}
\caption{Distribution of $p_r|_{ESG}$ for RL, $\text{MVO}_{\text{Exact}}$, $\text{MVO}_{\text{Relaxed}}$ for ESG-responsible utility functions in evaluation period.}
\label{fig:results_p_r}
\end{figure*}

We include cumulative financial returns for all strategies in Figure \ref{fig:results_all}. Additionally, we provide annualized financial returns in Table \ref{tab:results_fin} and average responsibility performance (average $p_r$) in Table \ref{tab:results_resp}. 

As it can be seen in Figure \ref{fig:results_all}, within RL results, multiplicative utility functions outperform additive utility functions in general for both $u_{\text{Sharpe}}$ and $u_{\text{Sortino}}$ financial objectives (in all cases except one). This highlights the value of multiplicative utility functions for incorporating responsibility objectives, validating our motivation to study Deep RL policies for responsible portfolio allocation.

With regards to financial performance, annualized financial returns for RL policies are consistent across utility functions evaluated, as financial returns do not vary significantly (from $9.9 \% $ to $12.53 \% $), whereas $\text{MVO}_{\text{Exact}}$ and $\text{MVO}_{\text{Relaxed}}$ exhibit significantly more variation (from $0.65 \%$ to $22.77 \%$ and from $-6.86 \%$ to $25.08 \%$ respectively). This increased variability in performance for MVO approaches is similar to the patterns observed by \citet{sood2023deep} when comparing Deep RL and MVO for portfolio optimization. In addition, we also note how the results in Table \ref{tab:max_drawdown} show that RL performs better in terms of maximum drawdown (computed across all utility functions) than $\text{MVO}_{\text{Exact}}$ and $\text{MVO}_{\text{Relaxed}}$.

With regards to responsibility performance, average $p_r$ is also not as varied across RL results, with the performance ratio for RL not being far from the benchmark performance of a uniform-allocation portfolio ($p_r$ values range from $-0.16 \%$ to $0.70 \%$). Alternatively, $\text{MVO}_{\text{Exact}}$ and $\text{MVO}_{\text{Relaxed}}$ provide better average responsibility performance, albeit with more variation ($p_r$ values range from $0.53 \%$ to $5.25 \%$ and from $0.04 \%$ to $13.35 \%$ respectively). Additionally, as it can be seen in Figure \ref{fig:results_p_r}, the distribution of daily $p_r$ during the evaluation period is more concentrated for RL strategies compared to $\text{MVO}_{\text{Exact}}$ and $\text{MVO}_{\text{Relaxed}}$ for both $u_{\text{Sharpe}}$ and $u_{\text{Sortino}}$ objectives.

Moreover, when comparing the financial performance across various responsibility objectives shown in Table \ref{tab:results_fin}, we highlight an interesting phenomenon observed between ESG objectives and E,S,G objectives: the financial performance of RL is generally above that of MVO approaches for responsibility objectives $u_E$, $u_S$, $u_G$ for both $u_{\text{Sharpe}}$ and $u_{\text{Sortino}}$ financial objectives. However, this trend reverses for responsibility objective $u_{ESG}$. Whilst we have not been able to identify a unique reason to explain this, we believe this is likely a combination of RL underperforming and MVO overperforming for this case. This could be partly related to the fact that, at least for the provider we have used, ESG scores reported by the provider are not an average of E, S, and G scores reported at any given time. We leave it for further work to study whether this phenomenon occurs for other ESG data providers, as this might depend on the specific recipes used by a provider to generate their scores. Variability of results depending on data provider has been previously highlighted as a limitation of portfolio optimization involving ESG scores \cite{henriksson2019integrating}. 
%When considering investor profiles for ESG portfolio optimization, \citet{pedersen2021responsible} suggest the following categorization: 
%\begin{itemize}
%    \item Type-U investors are ESG-unaware, i.e. they do not use ESG information at all and have no preference for ESG
%    \item Type-A investors are ESG-aware, i.e. they exploit ESG information but they do not enjoy ESG utility
%    \item Type-M are ESG-motivated, i.e. they use ESG information and have a preference for high average ESG score
%\end{itemize}

\begin{table}[]
\centering
\caption{Maximum drawdown across all utility functions for RL, $\text{MVO}_{\text{Exact}}$, $\text{MVO}_{\text{Relaxed}}$.}
\label{tab:max_drawdown}
\begin{tabular}{@{}cccc@{}}
\toprule
             & RL        & $\text{MVO}_{\text{Exact}}$ & $\text{MVO}_{\text{Relaxed}}$ \\ \midrule
Max Drawdown & -32.63 \% & -33.71 \%                   & -43.77 \%                     \\ \bottomrule
\end{tabular}
\end{table}

\section{Conclusion}
In this work we present a novel comparison between Deep RL and MVO strategies for responsible portfolio optimization, using various combinations of financial and ESG objectives. We formulate two modifications to mean-variance approaches which are similar to those proposed in previous work, and additionally we expand previous formulations for portfolio optimization using RL with responsibility objectives which are incorporated into the utility (reward) function in additive or multiplicative ways. We provide comprehensive comparisons between Deep RL and MVO strategies from the perspective of financial and responsibility objectives. Our results show that Deep RL is a competitive alternative to modified MVO approaches for responsible portfolio optimization, whilst exhibiting less variability. 

As future work, we would like to incorporate additional elements to the RL formulations such as nonlinear transaction costs, cardinality constraints on the portfolio allocation, conditional value-at-risk, and other factors which are not suitable for mean-variance approaches due to convexity requirements. Evaluating such strategies in benchmarks that incorporate these more realistic limitations could highlight the potential of RL approaches for non-convex responsible portfolio optimization. 
We hope this work contributes towards the development of robust methodologies for responsible investing and its increased adoption, in order to contribute to positive societal impact of Artificial Intelligence in Finance. 

\appendix

\section{Acknowledgments}
% Acknowledgements here.
%\iffalse
\paragraph{Disclaimer.}
This paper was prepared for informational purposes by
the Artificial Intelligence Research group of JPMorgan Chase \& Co. and its affiliates (``JP Morgan''),
and is not a product of the Research Department of JP Morgan.
JP Morgan makes no representation and warranty whatsoever and disclaims all liability,
for the completeness, accuracy or reliability of the information contained herein.
This document is not intended as investment research or investment advice, or a recommendation,
offer or solicitation for the purchase or sale of any security, financial instrument, financial product or service,
or to be used in any way for evaluating the merits of participating in any transaction,
and shall not constitute a solicitation under any jurisdiction or to any person,
if such solicitation under such jurisdiction or to such person would be unlawful.
%\fi

%\newpage

\bibliography{aaai24}

\end{document}